\documentclass[lettersize,journal]{IEEEtran}
\usepackage{amsmath,amsfonts}
\usepackage{array}
\usepackage[caption=false,font=footnotesize,labelfont=rm,textfont=rm]{subfig}
\usepackage{textcomp}
\usepackage{stfloats}
\usepackage{url}
\usepackage{verbatim}
\usepackage{graphicx}
\usepackage{cite}
\RequirePackage{titlesec}
\RequirePackage{titletoc}
\usepackage{amssymb} 
\usepackage{algorithm}  
\usepackage{algorithmicx}
\usepackage{algpseudocode}

\usepackage{booktabs}
\usepackage{orcidlink}
\usepackage{tikz}
\usepackage{bbm}

\hypersetup{
	colorlinks=true,
	linkcolor=blue,
	citecolor=blue
}

\newtheorem{theorem}{\bf Theorem}
\newtheorem{lemma}{\bf Lemma}
\newtheorem{corollary}{\bf Corollary}

\begin{document}
	
	\title{Dropout Strategy in Reinforcement Learning: Limiting the Surrogate Objective Variance in Policy Optimization Methods}
	
	\author{Zhengpeng Xie$^{\orcidlink{0009-0009-4159-5438}}$, Changdong Yu$^{\orcidlink{0000-0002-5759-4589}}$, Weizheng Qiao$^{\orcidlink{0000-0002-6770-4497}}$
		\thanks{Zhengpeng Xie is with the School of Mathematical Sciences, Chongqing Normal University, Chongqing 401331, China (e-mail: xzp3464479031@126.com).}
		\thanks{Changdong Yu (Corresponding author) is with the College of Artificial Intelligence, Dalian Maritime University, Dalian 116026, China (e-mail: ycd\_darren@163.com).}
		\thanks{Weizheng Qiao is with the College of Information Engineering, Minzu University of China, Beijing 100081, China (e-mail: 202101064@muc.edu.cn).}
		}
	
	
	\IEEEpubid{}
	
	\maketitle
	
	\begin{abstract}
		Policy-based reinforcement learning algorithms are widely used in various fields. Among them, mainstream policy optimization algorithms such as TRPO and PPO introduce importance sampling into policy iteration, which allows the reuse of historical data. However, this can also lead to a high variance of the surrogate objective and indirectly affects the stability and convergence of the algorithm. In this paper, we first derived an upper bound of the surrogate objective variance, which can grow quadratically with the increase of the surrogate objective. Next, we proposed the dropout technique to avoid the excessive increase of the surrogate objective variance caused by importance sampling. Then, we introduced a general reinforcement learning framework applicable to mainstream policy optimization methods, and applied the dropout technique to the PPO algorithm to obtain the D-PPO variant. Finally, we conduct comparative experiments between D-PPO and PPO algorithms in the Atari 2600 environment, and the results show that D-PPO achieved significant performance improvements compared to PPO, and effectively limited the excessive increase of the surrogate objective variance during training.
	\end{abstract}
	
	\begin{IEEEkeywords}
		Deep Reinforcement Learning, policy optimization, importance sampling, actor-critic, proximal policy optimization, surrogate objective variance, dropout strategy
	\end{IEEEkeywords}
	
	\section{Introduction}
	\IEEEPARstart{D}{eep} Reinforcement Learning (DRL) is a machine learning approach that combines deep learning and reinforcement learning to address end-to-end sequential decision-making problems. In DRL, an agent interacts with the environment and explores through trial and error to learn an optimal policy. In recent years, a number of DRL algorithms have been widely used in various fields, including board games\cite{AlphaGo, AlphaGoZero, AlphaZero}, video games\cite{DQN, Agent57, starcraft, king-of-glory}, autonomous driving\cite{autonomous-driving, autonomous-driving-2}, intelligent control\cite{robot-control, robotic-arm-control, control}, and so on. DRL has emerged as one of the hottest research topics in artificial intelligence.
	
	During the development of DRL, scholars have proposed and improved many representative methods, which can be summarized into two categories: 1) value-based and 2) policy-based DRL algorithms. Value-based DRL algorithms originated from Deep Q-Networks (DQN)\cite{DQN}, which approximates the action value function $Q(s,a)$ using a deep neural network and updates the network through the Bellman equation using historical data. Subsequent scholars have made a series of improvements to DQN, such as Schaul \textit{et al.}\cite{PER} proposing the prioritized experience replay technique, which prioritizes the use of data with larger TD errors for network updates, improving the learning efficiency. Hasselt \textit{et al.}\cite{double-dqn} proposed a Double Q-learning algorithm that mitigates the problem of overestimating the action value function $Q(s,a)$ in DQN by introducing a target network in the update process. Wang \textit{et al.}\cite{dueling-dqn} proposed a Dueling Network that decomposes the action value function $Q(s,a)$ into a state value function $V(s)$ and an advantage function $A(s,a)$, with both of them shares the same convolutional layer during training. The Dueling Network can better estimate the contribution of different actions to the state, improving the learning efficiency of DQN. Bellemare \textit{et al.}\cite{distributional-dqn} modeled the distribution of the action-state value function $Q(s,a)$ in DQN to avoid losing information about the distribution of it during the training process, thereby improving the performance of the DQN algorithm. Fortunato \textit{et al.}\cite{NoisyNet} proposed NoisyNet, which adds random noise to the fully connected layers of the network to enhance the exploration and robustness of the model. Finally, Hessel \textit{et al.}\cite{rainbow-dqn} integrated all the excellent variants of DQN and used a multi-step learning approach to calculate the error, resulting in the Rainbow DQN, which achieved state-of-the-art performance.
	
	Unlike value-based reinforcement learning methods, policy-based reinforcement learning methods directly learn a policy network that outputs a probability distribution of actions based on the input state, and randomly samples an action from it. Therefore, it can effectively solve the problem of high-dimensional continuous action spaces. The policy-based reinforcement learning algorithm originated from the Reinforce algorithm proposed by Sutton \textit{et al.}\cite{pg}, which uses Monte Carlo (MC) method to approximate the policy gradient estimation. Subsequently, many policy optimization methods were proposed\cite{npo,ppo_5}, with TRPO and PPO being the most representative ones. The Trust Region Policy Optimization (TRPO) algorithm is proposed by Schulman \textit{et al.}\cite{trpo}, which introduces the trust region approach in optimization theory to policy optimization, representing the policy update process as a constrained optimization problem. By limiting the KL divergence between the old and new policies, the TRPO algorithm limits the change in policy parameters in the update process. They also provided theoretical guarantees that TRPO algorithm has monotonic improvement performance, which makes TRPO algorithm more robust to hyperparameter settings and more stable than traditional policy gradient methods. However, TRPO algorithm needs to solve a constrained optimization problem in each round of update, which also leads to a significant computational overhead of TRPO algorithm, making it unsuitable for solving large-scale reinforcement learning problems. Schulman \textit{et al.}\cite{ppo} proposed proximal policy optimization (PPO) algorithm, PPO algorithm has two main variants, namely PPO-CLIP and PPO-PENALTY, where PPO-PENALTY also uses the KL divergence between old and new policies as a constraint, but treats it as a penalty term in the objective function rather than a hard constraint, and automatically adjusts the penalty coefficient during policy update to ensure constraint satisfaction; PPO-CLIP does not use KL divergence, but introduces a special ratio clipping function, which limits the ratio of output action probabilities between old and new policies through a clipping function, thus implicitly ensuring that the algorithm satisfies the constraints of old and new policies during the update process. PPO-CLIP algorithm is widely used by scholars in many reinforcement learning environments due to its concise form, ease of programming implementation, and superior performance. Many subsequent studies have discussed whether the ratio clipping function in the PPO algorithm can effectively guarantee the constraint of the trust region\cite{clip1,clip2,clip3,clip4,clip5,clip6,clip7}. However, at present, the equivalence between the ratio clipping function and the confidence domain constraint is not clear. Although TRPO and PPO algorithms are widely used in many reinforcement learning environments, importance sampling can cause their surrogate objective variance to become very large during training, which is an urgent problem that needs to be addressed.
	
	The process of policy optimization relies on the estimation of policy gradients, and its accuracy depends on the variance and bias caused by the estimated policy. A common approach to reducing variance is to add a value function that is independent of actions as a baseline to the policy update process. Generally, a state value function is used as the baseline, which naturally leads to the Actor-Critic (AC)\cite{ac} architecture. The introduction of a baseline can significantly reduce the variance of the policy gradient estimate. Schulman \textit{et al.}\cite{gae} proposed a technique called Generalized Advantage Estimation (GAE), which is an extension of the advantage function. GAE uses an exponential weighted estimate similar to the advantage function to reduce the bias of the policy gradient estimate. However, the above research is conducted for policy gradients, and there is a lack of systematic analysis and discussion on the variance of the objective function.

	In summary, we focus on addressing the issue of excessive growth in the variance of the surrogate objective during the iterative process of introducing importance sampling strategies. Our main contributions are summarized as follows.
	\begin{enumerate}
		\item For the iterative process of introducing the importance sampling strategy, we derive the variance upper bound of its surrogate objective, and show that this upper bound approximately grows quadratically with the increase of the surrogate objective. To the best of our knowledge, we are the first to provide this upper bound.
		\item A general reinforcement learning framework for traditional policy optimization methods is proposed, and the mathematical formalization of the dropout strategy is given.
		\item Two feasible dropout strategies are proposed, and the feasibility of the proposed dropout strategy is explained based on the theoretical results of the surrogate objective variance.
		\item Introducing the dropout technique into the PPO algorithm to obtain the D-PPO algorithm. A series of comparative experiments with the PPO algorithm in the Atari 2600 environment show that the performance of the D-PPO algorithm is superior to that of PPO, and it has a lower variance of the surrogate objective.
	\end{enumerate}
	
	The remainder of this article is organized as follows: Section \ref{Related Work} introduces the policy gradient and related work, including TRPO and PPO algorithms. Section \ref{Dropout Strategy in Policy Optimization} introduces the main theoretical results, dropout technique, dropout strategy framework, and pseudo-code of D-PPO algorithm. Section \ref{Experiments} presents the comparative experiments between D-PPO and PPO algorithms and the hyperparameter analysis of D-PPO algorithm. Section \ref{Conclusion} summarizes this article and presents the conclusion.

	\section{Related Work}\label{Related Work}
	In this section, we will briefly introduce some basic concepts of reinforcement learning and two representative policy optimization methods.
	
	\subsection{Policy Gradient}
	Reinforcement learning is generally defined by a tuple $(\mathcal{S},\mathcal{A},r,\mathcal{P},\rho_0,\gamma)$, where $\mathcal{S}$ and $\mathcal{A}$ represent the state space and action space, $r:\mathcal{S}\times\mathcal{A}\rightarrow\mathbb{R}$ is the reward function, $\mathcal{P}:\mathcal{S}\times\mathcal{A}\times\mathcal{S}\rightarrow[0,1]$ is the probability distribution of the state transition function, $\rho_0$ is the initial state distribution, and $\gamma\in[0,1]$ is the discount factor. Starting from the initial state, for each time step, the agent receives the current state $s_t$, takes the action $a_t$, obtains the reward $r_t$ from the environment feedback, and obtains next state $s_{t+1}$ until entering the terminal state. The action value function and state value function are defined as
	\begin{equation*}
		Q^{\pi}(s_t,a_t):=\mathbb{E}_{\mathcal{S}_{t+1},\mathcal{A}_{t+1}}\left[\sum_{i=t}^{T}\gamma^{i-t}r_i\Bigg|S_t=s_t,A_t=a_t\right],
	\end{equation*}
	\begin{equation*}
		V^{\pi}(s_t):=\mathbb{E}_{\mathcal{S}_{t+1},\mathcal{A}_t}\left[\sum_{i=t}^{T}\gamma^{i-t}r_i\Bigg|S_t=s_t\right],
	\end{equation*}
	where $\mathcal{S}_{t}=\left\{S_{t},\dots,S_T\right\},\mathcal{A}_{t}=\left\{A_{t}\dots,A_T\right\}$.

	Policy-based reinforcement learning algorithms often require the estimation of Policy Gradient (PG), so the derivation of policy gradient is briefly introduced below. Consider the trajectory generated by an agent starting from an initial state and interacting with the environment for one full episode
	\begin{equation}
		\tau=(s_1,a_1,r_1,\dots,s_{T-1},a_{T-1},r_{T-1},s_T).
	\end{equation}
	The goal of reinforcement learning is to maximize the expected return $R(\tau)=\sum_{i=1}^{T}\gamma^{i-1}r_i$, so the expected $R(\tau)$ for all possible trajectories is
	\begin{equation}
		J(\theta)=\mathbb{E}_{\tau\sim p(\cdot)}\left[R(\tau)\right]=\sum_{\tau}R(\tau)\cdot p(\tau),
	\end{equation}
	where $p(\tau)=\rho_0(s_1)\cdot\prod_{t=1}^{T-1}\pi_{\theta}(a_t|s_t)\cdot \mathcal{P}(s_{t+1}|s_t,a_t)$, and $\pi_{\theta}$ is the parameterized policy network. The gradient of the objective function $J(\theta)$ with respect to parameters $\theta$ is obtained by
	\begin{equation}\label{pg}
		\begin{split}
			\nabla J(\theta)&=\sum_{\tau}R(\tau)\cdot\nabla p(\tau)=\sum_{\tau}R(\tau)\cdot\nabla \log p(\tau)\cdot p(\tau) \\
			&=\mathbb{E}_{\tau\sim p(\cdot)}\left[R(\tau)\cdot\nabla\log p(\tau)\right] \\
			&=\mathbb{E}_{\tau\sim p(\cdot)}\Bigg\{R(\tau)\cdot\nabla\Bigg[\log\rho_0(s_1)+\sum_{t=1}^{T-1}\log\pi_{\theta}(a_t|s_t)\\
			&\qquad\qquad\quad+\log\mathcal{P}(s_{t+1}|s_t,a_t)\Bigg]\Bigg\}  \\
			&=\mathbb{E}_{\tau\sim p(\cdot)}\left[\sum_{t=1}^{T-1}R(\tau)\cdot\nabla\log\pi_{\theta}(a_t|s_t)\right]\\
			&\approx\frac{1}{N}\sum_{n=1}^{N}\sum_{t=1}^{T_n-1}R(\tau^n)\cdot\nabla\log\pi_{\theta}(a_t^n|s_t^n). \\
		\end{split}
	\end{equation}

	We derived the basic form of the policy gradient, so that the agent can approximate the policy gradient based on the current policy network $\pi_{\theta}$ and $N$ episodes of interaction with the environment, which is called Monte Carlo method.
	
	\subsection{Trust Region Policy Optimization}
	If we adopt the equation $(\ref{pg})$ to estimate the policy gradient, the data collected by the current policy network can only be used to update the policy network parameters $\theta$ once, which will lead to inefficient sample utilization. To enable the agent to reuse the data generated by the old policy and improve sample utilization efficiency, we can introduce importance sampling, that is, 
	\begin{equation}\label{importance sampling}
		\mathbb{E}_{x\sim p(x)}f(x)=\mathbb{E}_{x\sim q(x)}\frac{p(x)}{q(x)}\cdot f(x).
	\end{equation}
	
	The condition for equation $(\ref{importance sampling})$ to hold in sampling estimation is that the difference between probability distributions $p$ and $q$ cannot be too large. Therefore, Schulman \textit{et al.}\cite{trpo} introduces the KL divergence between the current policy $\pi_{\theta}$ and the old policy $\pi_{\theta_{\rm old}}$ as a constraint in the iterative update process of reinforcement learning, and proposes the Trust Region Policy Optimization (TRPO) algorithm, which is
	\begin{equation}
		\begin{split}			
			\max_{\theta}\enspace &\mathbb{E}_{(s_t,a_t)\sim\pi_{\theta_{\rm old}}}\left[\frac{\pi_{\theta}(a_t|s_t)}{\pi_{\theta_{\rm old}}(a_t|s_t)}\cdot \hat{A}(s_t,a_t)\right] \\
			\text{s.t.}\hspace{2.5pt}\enspace &\mathbb{E}\left[\text{KL}(\pi_{\theta},\pi_{\theta_{\rm old}})\right]\leq\delta.      \\
		\end{split}
	\end{equation}
	Where $\hat{A}(s_t,a_t)$ is the estimated value of the advantage function at time $t$.

	\subsection{Proximal Policy Optimization}
	Although the TRPO algorithm theoretically ensures monotonic improvement of the policy, it requires solving a constrained optimization problem in each iteration, which results in low computational efficiency and makes it difficult to apply to large-scale reinforcement learning tasks. To address this issue, Schulman \textit{et al.}\cite{ppo} proposed the PPO algorithm, which has two main variants: 1) one of them is called PPO-CLIP, that is,  
	\begin{equation}\label{ppo-clip}
		\begin{split}
			\max_{\theta}\enspace&\mathbb{E}_{(s_t,a_t)\sim\pi_{\theta_{\text{old}}}}\Bigg\{\min\Bigg[\frac{\pi_{\theta}(a_t|s_t)}{\pi_{\theta_{\text{old}}}(a_t|s_t)}\cdot\hat{A}(s_t,a_t), \\
			&\text{clip}\left(\frac{\pi_{\theta}(a_t|s_t)}{\pi_{\theta_{\text{old}}}(a_t|s_t)},1-\epsilon,1+\epsilon\right)\cdot\hat{A}(s_t,a_t)\Bigg]\Bigg\}; \\
		\end{split}
	\end{equation}
	2) another one is called PPO-PENALTY, which can be formalized as
	\begin{equation}\label{ppo-penalty}
		\max_{\theta}\mathop{\mathbb{E}}_{(s_t,a_t)\sim\pi_{\theta_{\rm old}}}\left\{\frac{\pi_{\theta}(a_t|s_t)}{\pi_{\theta_{\text{old}}}(a_t|s_t)}\hat{A}(s_t,a_t)-\beta\left[\text{KL}(\pi_{\theta},\pi_{\theta_{\rm old}})\right]\right\},
	\end{equation}
	where $\beta$ is adaptive KL penalty coefficient.

	\section{Dropout Strategy in Policy Optimization}\label{Dropout Strategy in Policy Optimization}
	This section presents the main theoretical results for the surrogate objective variance, deriving its upper bound, and propose a dropout strategy, including its abstract representation.
	
	\subsection{Main Results}
	\textbf{Symbol Description.} We adopt
	\begin{equation}
		\mathfrak{O}_{\theta_{\rm old}}^{\theta}(s,a)\triangleq\frac{\pi_{\theta}(a|s)}{\pi_{\theta_{\rm old}}(a|s)}\cdot A^{\pi_{\theta_{\rm old}}}(s,a)
	\end{equation}
	to denote the surrogate objective, and use $\mathbb{P}_{\theta_{\rm old}}(s)$ to represent the probability of state $s$ occurring under the current policy network parameters $\theta_{\rm old}$, which is \textbf{not computable}. Thus, we have
	\begin{equation}\label{p(s,a)}
		\mathbb{P}_{\theta_{\rm old}}(s,a)=\mathbb{P}_{\theta_{\rm old}}(s)\cdot\pi_{\theta_{\rm old}}(a|s)
	\end{equation}
	to represent the probability of tuple $(s,a)$ occurrence.
	
	\begin{lemma}\label{lemma_1}
		The expectation of the square of the surrogate objective can be written as
		\begin{equation*}
			\mathop{\mathbb{E}}_{(s,a)\sim\pi_{\theta_{\rm old}}}\left\{\left[\mathfrak{O}_{\theta_{\rm old}}^{\theta}(s,a)\right]^2\right\}=\mathop{\mathbb{E}}_{(s_i,a_i)\sim\pi_{\theta_{\rm old}}}\left\{\left[\mathfrak{O}_{\theta_{\rm old}}^{\theta}(s_i,a_i)\right]^2\right\}.
		\end{equation*}
	\end{lemma}
	
	\begin{IEEEproof}
		We just added an index to data $(s,a)$ as a preparation, this will not change the expectation.
	\end{IEEEproof}
	
	\begin{lemma}\label{lemma_2}
		The square of the expectation of the surrogate objective can be written as
		\begin{equation*}
			\begin{split}
				&\left\{\mathop{\mathbb{E}}_{(s,a)\sim\pi_{\theta_{\rm old}}}\left[\mathfrak{O}_{\theta_{\rm old}}^{\theta}(s,a)\right]\right\}^2\\
				=&\mathop{\mathbb{E}}_{(s_i,a_i)\sim\pi_{\theta_{\rm old}}}\Bigg\{\mathbb{P}_{\theta_{\rm old}}(s_i)\cdot\pi_{\theta_{\rm old}}(a_i|s_i)\cdot\left[\mathfrak{O}_{\theta_{\rm old}}^{\theta}(s_i,a_i)\right]^2\\
				+&\mathop{\mathbb{E}}_{(s_j,a_j)\sim\pi_{\theta_{\rm old}}\atop j\neq i}\left[\mathfrak{O}_{\theta_{\rm old}}^{\theta}(s_i,a_i)\cdot\mathfrak{O}_{\theta_{\rm old}}^{\theta}(s_j,a_j)\right]\Bigg\}.\\
			\end{split}
		\end{equation*}
	\end{lemma}

	\begin{IEEEproof}
		According to the definition of expectation and equation $(\ref{p(s,a)})$, we have
	\begin{equation*}
		\begin{split}
			&\left\{\mathop{\mathbb{E}}_{(s,a)\sim\pi_{\theta_{\rm old}}}\left[\mathfrak{O}_{\theta_{\rm old}}^{\theta}(s,a)\right]\right\}^2\\
			=&\left[\sum_{(s,a)\sim\pi_{\theta_{\rm old}}}\mathbb{P}_{\theta_{\rm old}}(s,a)\cdot\frac{\pi_{\theta}(a|s)}{\pi_{\theta_{\rm old}}(a|s)}\cdot A^{\pi_{\theta_{\rm old}}}(s,a)\right]^2\\
			=&\left[\sum_{(s,a)\sim\pi_{\theta_{\rm old}}}\mathbb{P}_{\theta_{\rm old}}(s)\cdot\pi_{\theta}(a|s)\cdot A^{\pi_{\theta_{\rm old}}}(s,a)\right]^2\\
			=&{\sum_{(s_i,a_i)\sim\pi_{\theta_{\rm old}}}}\mathbb{P}_{\theta_{\rm old}}(s_i)^2\cdot\pi_{\theta}(a_i|s_i)^2\cdot A^{\pi_{\theta_{\rm old}}}(s_i,a_i)^2+\\
			&{\sum_{(s_i,a_i)\sim\pi_{\theta_{\rm old}}}}\mathbb{P}_{\theta_{\rm old}}(s_i)\cdot\pi_{\theta}(a_i|s_i)\cdot A^{\pi_{\theta_{\rm old}}}(s_i,a_i)\cdot\\
			&{\sum_{\substack{(s_j,a_j)\sim\pi_{\theta_{\text{old}}}\\j\neq i}}}\mathbb{P}_{\theta_{\rm old}}(s_j)\cdot\pi_{\theta}(a_j|s_j)\cdot A^{\pi_{\theta_{\rm old}}}(s_j,a_j)\\
			=&{\sum_{(s_i,a_i)\sim\pi_{\theta_{\rm old}}}}\mathbb{P}_{\theta_{\rm old}}(s_i,a_i)^2\cdot\frac{\pi_{\theta}(a_i|s_i)^2}{\pi_{\theta_{\rm old}}(a_i|s_i)^2}\cdot A^{\pi_{\theta_{\rm old}}}(s_i,a_i)^2+\\
			&{\sum_{(s_i,a_i)\sim\pi_{\theta_{\rm old}}}}\mathbb{P}_{\theta_{\rm old}}(s_i,a_i)\cdot\frac{\pi_{\theta}(a_i|s_i)}{\pi_{\theta_{\rm old}}(a_i|s_i)}\cdot A^{\pi_{\theta_{\rm old}}}(s_i,a_i)\cdot\\
			&{\sum_{\substack{(s_j,a_j)\sim\pi_{\theta_{\text{old}}}\\j\neq i}}}\mathbb{P}_{\theta_{\rm old}}(s_j,a_j)\cdot\frac{\pi_{\theta}(a_j|s_j)}{\pi_{\theta_{\rm old}}(a_j|s_j)}\cdot A^{\pi_{\theta_{\rm old}}}(s_j,a_j)\\
			=&{\mathop{\mathbb{E}}_{(s_i,a_i)\sim\pi_{\theta_{\rm old}}}}\mathbb{P}_{\theta_{\rm old}}(s_i,a_i)\cdot\left[\mathfrak{O}_{\theta_{\rm old}}^{\theta}(s_i,a_i)\right]^2+\\
			&{\mathop{\mathbb{E}}_{(s_i,a_i)\sim\pi_{\theta_{\rm old}}}\mathop{\mathbb{E}}_{(s_j,a_j)\sim\pi_{\theta_{\rm old}}\atop j\neq i}}\left[\mathfrak{O}_{\theta_{\rm old}}^{\theta}(s_i,a_i)\cdot\mathfrak{O}_{\theta_{\rm old}}^{\theta}(s_j,a_j)\right].\\
		\end{split}
	\end{equation*}
		Hence, Lemma \ref{lemma_2} is proved.
	\end{IEEEproof}

	\begin{theorem}\label{theorem_1}
		When introducing importance sampling, the variance of the surrogate objective $\mathbb{E}_{(s,a)\sim\pi_{\theta_{\rm old}}}\left[\mathfrak{O}_{\theta_{\rm old}}^{\theta}(s,a)\right]$ can be written as
		\begin{equation*}
			\begin{split}
				&\sigma_{\theta_{\rm old}}(\theta)=\\
				&\mathop{\mathbb{E}}_{(s_i,a_i)\sim\pi_{\theta_{\rm old}}}\left\{\xi-\mathop{\mathbb{E}}_{(s_j,a_j)\sim\pi_{\theta_{\rm old}}\atop j\neq i}\left[\mathfrak{O}_{\theta_{\rm old}}^{\theta}(s_i,a_i)\cdot\mathfrak{O}_{\theta_{\rm old}}^{\theta}(s_j,a_j)\right]\right\},\\
			\end{split}
		\end{equation*}
		where $\xi=\left[1-\mathbb{P}_{\theta_{\rm old}}(s_i)\cdot\pi_{\theta_{\rm old}}(a_i|s_i)\right]\cdot \left[\mathfrak{O}_{\theta_{\rm old}}^{\theta}(s_i,a_i)\right]^2$.
	\end{theorem}
	
	\begin{IEEEproof}
		According to
		\begin{equation*}
		\begin{split}
			&\sigma_{\theta_{\rm old}}(\theta)\\
			=&\mathop{\mathbb{E}}_{(s,a)\sim\pi_{\theta_{\rm old}}}\left\{\left[\mathfrak{O}_{\theta_{\rm old}}^{\theta}(s,a)\right]^2\right\}-\left\{\mathop{\mathbb{E}}_{(s,a)\sim\pi_{\theta_{\rm old}}}\left[\mathfrak{O}_{\theta_{\rm old}}^{\theta}(s,a)\right]\right\}^2\\
		\end{split}
		\end{equation*}
		and Lemma \ref{lemma_1}-\ref{lemma_2}, Theorem \ref{theorem_1} is proved.
	\end{IEEEproof}

	\begin{corollary}\label{corollary_1}
		When introducing importance sampling, the variance of the surrogate objective $\mathbb{E}_{(s,a)\sim\pi_{\theta_{\rm old}}}\left[\mathfrak{O}_{\theta_{\rm old}}^{\theta}(s,a)\right]$ is bounded by
		\begin{equation*}
			\begin{split}
				\sigma_{\theta_{\rm old}}(\theta)\leq&\mathop{\mathbb{E}}_{(s_i,a_i)\sim\pi_{\theta_{\rm old}}}\Bigg\{\left[\mathfrak{O}_{\theta_{\rm old}}^{\theta}(s_i,a_i)\right]^2-\\
				&\mathop{\mathbb{E}}_{(s_j,a_j)\sim\pi_{\theta_{\rm old}}\atop j\neq i}\left[\mathfrak{O}_{\theta_{\rm old}}^{\theta}(s_i,a_i)\cdot\mathfrak{O}_{\theta_{\rm old}}^{\theta}(s_j,a_j)\right]\Bigg\}.\\
			\end{split}
		\end{equation*}
	\end{corollary}

	\begin{IEEEproof}
		According to
		\begin{equation*}
		\begin{split}
			\xi&=\left[1-\mathbb{P}_{\theta_{\rm old}}(s_i)\cdot\pi_{\theta_{\rm old}}(a_i|s_i)\right]\cdot \left[\mathfrak{O}_{\theta_{\rm old}}^{\theta}(s_i,a_i)\right]^2\\
			&\leq\left[\mathfrak{O}_{\theta_{\rm old}}^{\theta}(s_i,a_i)\right]^2\\
		\end{split}
		\end{equation*}
		in Theorem \ref{theorem_1}, we have the Corollary \ref{corollary_1}, which means removing the uncomputable $\mathbb{P}_{\theta_{\rm old}}(s_i)$.
	\end{IEEEproof}
	
	\textbf{Explanation.}
	It is clear from Corollary \ref{corollary_1} that the upper bound on the variance of the surrogate objective is mainly determined by two terms: 1) one of which is the square of the surrogate objective, which means that increasing the objective function will inevitably lead to a quadratic increase in the variance with respect to it; 2) the other is
	\begin{equation}\label{expectation}
		\mathop{\mathbb{E}}_{(s_j,a_j)\sim\pi_{\theta_{\rm old}}\atop j\neq i}\left[\mathfrak{O}_{\theta_{\rm old}}^{\theta}(s_i,a_i)\cdot\mathfrak{O}_{\theta_{\rm old}}^{\theta}(s_j,a_j)\right],
	\end{equation}
	which is subtracted from the square of the surrogate objective. Therefore, in order to reduce the variance of the surrogate objective, we mainly focus on adjusting this item from the perspective of training data.
	
	\subsection{Dropout Strategy and Formalization}\label{Dropout Strategy}
	Now suppose that the agent interacts with the environment to obtain training data\footnote{Here we ignore the terminal states.}
	\begin{equation}
		(s_1,a_1,r_1),(s_2,a_2,r_2),\dots,(s_{n},a_{n},r_{n}),
	\end{equation}
	for each data $(s_i,a_i,r_i)$, we denote its corresponding expectation $(\ref{expectation})$ as $\Delta_i$ and perform Monte Carlo approximation, which is denoted as
	\begin{equation}
		\Delta_i\approx\hat{\Delta}_i=\sum_{\substack{(s_j,a_j)\sim\pi_{\theta_{\text{old}}}\\j\neq i}}\left[\hat{\mathfrak{O}}_{\theta_{\rm old}}^{\theta}(s_i,a_i)\cdot\hat{\mathfrak{O}}_{\theta_{\rm old}}^{\theta}(s_j,a_j)\right],
	\end{equation}
	where $i=1,2,\dots,n$; $\hat{\mathfrak{O}}_{\theta_{\rm old}}^{\theta}(s,a)=\frac{\pi_{\theta}(a|s)}{\pi_{\theta_{\rm old}}(a|s)}\cdot\hat{A}^{\pi_{\theta_{\rm old}}}(s,a)$ and $\hat{A}^{\pi_{\theta_{\rm old}}}(s,a)$ is an estimate of the advantage $A^{\pi_{\theta_{\rm old}}}(s,a)$, using GAE\cite{gae} technique.
	
	At the code level, we implement parallel computation of $\hat{\Delta}_i$ through matrices, that is, 
	\begin{equation}
		\begin{bmatrix}
			\hat{\Delta}_1\\
			\hat{\Delta}_2\\
			\vdots\\
			\hat{\Delta}_n\\
		\end{bmatrix}=
		\begin{bmatrix} 
			\hat{\mathfrak{O}}_1&\hat{\mathfrak{O}}_1&\cdots&\hat{\mathfrak{O}}_1 \\
			\hat{\mathfrak{O}}_2&\hat{\mathfrak{O}}_2&\cdots & \hat{\mathfrak{O}}_2\\
			\vdots&\vdots& \ddots&\vdots \\
			\hat{\mathfrak{O}}_n&\hat{\mathfrak{O}}_n&\cdots&\hat{\mathfrak{O}}_n \\
		\end{bmatrix}\cdot
		\begin{bmatrix}
			\hat{\mathfrak{O}}_1\\
			\hat{\mathfrak{O}}_2\\
			\vdots\\
			\hat{\mathfrak{O}}_n\\
		\end{bmatrix}-
		\begin{bmatrix}
			\hat{\mathfrak{O}}_1^2\\
			\hat{\mathfrak{O}}_2^2\\
			\vdots\\
			\hat{\mathfrak{O}}_n^2\\
		\end{bmatrix},
	\end{equation}
	where $\hat{\mathfrak{O}}_i=\hat{\mathfrak{O}}_{\theta_{\rm old}}^{\theta}(s_i,a_i)$.
	
	So far, we have given an estimate of the second term $\hat{\Delta}_i$ in Corollary \ref{corollary_1}. Next, we will give a specific sample dropout strategy used in this paper. Before that, we would like to first introduce some abstract mathematical definitions.
	
	We define
	\begin{equation}
		\mathbb{D}_\phi^f(X):=\left\{x|x\in X,f(\phi(x))>0\right\},
	\end{equation}
	where $X=\left\{(s_i,a_i,r_i)\right\}_{i=1}^{n}$ is the training dataset, $\phi:\mathcal{S}\times\mathcal{A}\times\mathbb{R}\rightarrow\mathbb{R}$ represents a certain transformation applied to each data $x$ in the dataset $X$, and $f$ corresponds to a certain filtering rule for the original dataset $X$. Therefore, $\mathbb{D}_\phi^f$ is a formalization of the dropout strategy, which maps the original data $X$ to a subset of it, that means, $\mathbb{D}_\phi^f(X)\subset X$.
	
\begin{figure}[H]
		\centering
		\subfloat[]{
			\includegraphics[scale=0.9]{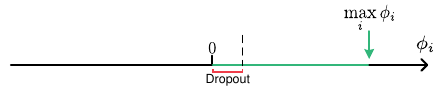}
		}
		
		\subfloat[]{
			\includegraphics[scale=0.9]{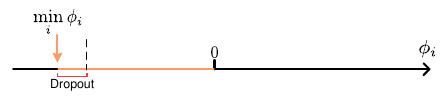}
		}
		\caption{Dropout strategy. (a) $\phi_i>0$. (b) $\phi_i<0$.}
		\label{phi}
	\end{figure}

	For example, $\phi(x_i)=\phi(s_i,a_i,r_i)$ denotes $\hat{\Delta}_i$ in this paper. As mentioned earlier, it can be seen from Corollary \ref{corollary_1} that $\Delta_i$ is subtracted from the previous term. In order to reduce the variance expectation of the surrogate objective, we want the value of $\hat{\Delta}_i$ to be as large as possible, whether it is positive or negative. This means that data $X$ is divided into two parts based on the sign of $\hat{\Delta}_i$, and for both of them, we choose to dropout the data $x_i$ corresponding to the relatively small $\hat{\Delta}_i$ to restrict the surrogate objective variance, as shown in Fig \ref{phi}.
	
	\begin{figure*}[!t]
		\centering
		\includegraphics[scale=0.75]{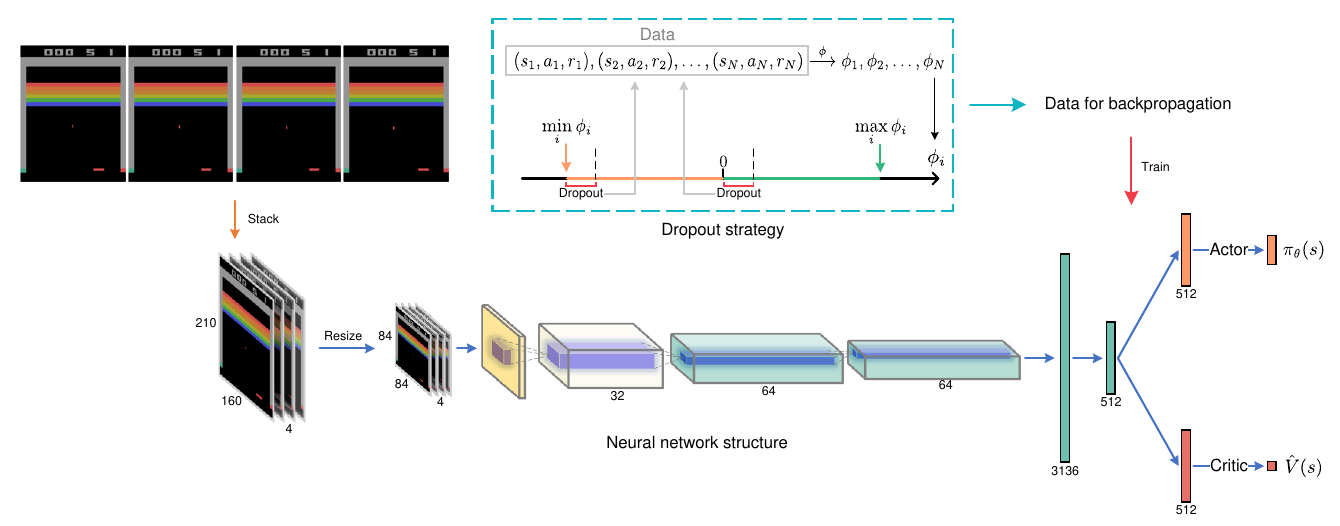}
		\caption{Dropout strategy and neural network structure.}
		\label{network structure}
	\end{figure*}
	
	Now suppose that the data $X$ is divided into two parts $X_{\phi}^+$ and $X_{\phi}^-$ according to the sign of $\hat{\Delta}_i$, that is,
	\begin{equation}
		\left\{ 
		\begin{split}
			& X_\phi^+=\left\{x|x\in X,\phi(x)\geq0\right\}; \\
			& X_\phi^-=\left\{x|x\in X,\phi(x)<0\right\}.    \\
		\end{split}\right.
	\end{equation}

	There are two ways to implement the dropout strategy: 1) one is setting a threshold $\delta^-$ and $\delta^+$ for $X_{\phi}^-$ and $X_{\phi}^+$, at this point, our dropout strategy is formalized as
	\begin{equation}
		\mathcal{D}(X)=\mathbb{D}_\phi^{x-\delta^-}(X_\phi^-)\cup\mathbb{D}_\phi^{x-\delta^+}(X_\phi^+).
	\end{equation}
	However, this way is too sensitive to the setting of the hyperparameters $\delta^-$ and $\delta^+$, and due to orders of magnitude and other factors, it may be difficult to select a pair of $\delta^-$ and $\delta^+$ that is applicable to any environment.
	
	2) The other is to fix the dropout ratio in the training dataset $X$, which introduces the hyperparameter $r\in[0,1]$. For a set of numbers, $M$, We define
	\begin{equation}
		M^{[r]}\triangleq\underset{m\in M}{\arg\min}\enspace\left|\frac{\left|\mathbb{D}_{\mathbbm{1}}^{m-x}\right|}{\left|M\right|}-r\right|,
	\end{equation}
	where $\mathbbm{1}(\cdot)$ represents the identity mapping, and $\left|\cdot\right|$ represents the absolute value or the number of elements in the set. Therefore, the dropout strategy can be formalized as
	\begin{equation}\label{dropout}
		\mathcal{D}(X)=\mathbb{D}_\phi^{x-[\phi(X_\phi^-)]^{[r]}}(X_\phi^-)\cup\mathbb{D}_\phi^{x-[\phi(X_\phi^+)]^{[r]}}(X_\phi^+),
	\end{equation}
	where $\phi(X)=\left\{\phi(x)|x\in X\right\}$.

	\subsection{Framework and Algorithm}
	Clearly, not all data can be effectively used to improve the performance of a policy network, especially in environments with sparse rewards\cite{sparse reward}. Most of the data collected by the agent through interaction with the environment is not directly helpful for improving its policy. Therefore, it is necessary to develop a dropout rule for sample data in specific situations to improve the learning efficiency of the algorithm.
	
	A specific dropout strategy is given in equation $(\ref{dropout})$, and we first present a more general dropout strategy framework, as shown in Algorithm \ref{Dropout strategy framework}. Where algorithm $\mathcal{A}$ can be any of the policy optimization algorithms, such as TRPO or PPO, which introduces importance sampling. When algorithm $\mathcal{A}$ is PPO algorithm and the drpout strategy is given by $(\ref{dropout})$, we obtain the pseudo-code of D-PPO algorithm in Algorithm \ref{D-PPO}.
	
	\begin{algorithm}[!t]
		\caption{Dropout strategy framework}\label{Dropout strategy framework}
		\begin{algorithmic}[1]
			\Require Policy and value network parameters $\theta,w$; policy optimization algorithm $\mathcal{A}$; dropout strategy $\mathcal{D}$
			\Ensure Optimal policy parameters $\theta^*$
			\While{not converged}
			\State Collect data $X=\left\{(s_i,a_i,r_i)\right\}_{i=1}^{N}$ using the current policy network $\pi_{\theta}$
			\For{each training  epoch}
			\State Use $X$ and $\mathcal{A}$ to update parameters:
			\begin{equation*}
				\theta,w\leftarrow\mathcal{A}(X;\theta,w)
			\end{equation*}
			\State Dropout strategy: $X\leftarrow\mathcal{D}(X)$
			\EndFor
			\EndWhile
		\end{algorithmic}  
	\end{algorithm}

	\section{Experiments}\label{Experiments}
	In this section, we first introduce the experimental environment, Atari 2600, as well as the structure of the policy network and value network, and the hyperparameter settings. Then we compare the performance of D-PPO and PPO algorithms in different environments. Finally, we analyze the impact of the hyperparameters of D-PPO on its performance.
	
	\subsection{Atari 2600}
\begin{figure}[!t]
		\centering
		\subfloat[]{
			\includegraphics[scale=0.4]{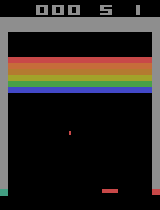}
		}
		\subfloat[]{
			\includegraphics[scale=0.4]{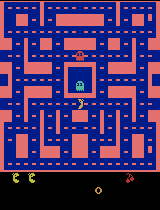}
		}
		\subfloat[]{
			\includegraphics[scale=0.4]{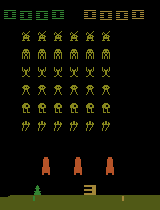}
		}
		\caption{Atari 2600 environments. (a) Breakout. (b) MsPacman. (c) SpaceInvaders.}
		\label{b and s}
	\end{figure}

	The Atari 2600 environment is a popular test platform for reinforcement learning, which was introduced by Atari in 1977 as a video game console. The environment contains various popular games such as Breakout, MsPacman and Space Invaders, as shown in Fig. \ref{b and s}. Since the introduction of the Deep Q-Networks (DQN) algorithm by Mnih \textit{et al.}\cite{DQN}, the Atari 2600 has become a standard environment for testing new reinforcement learning algorithms. Subsequently, OpenAI Gym packaged the Atari 2600 environment to create a more standardized interface and provided 57 Atari 2600 games as an environment. These games cover a variety of genres and difficulties, allowing researchers to experiment and compare different problems and algorithms.

	\begin{algorithm}[h]
		\caption{D-PPO}\label{D-PPO}
		\begin{algorithmic}[1]
			\Require Policy and value network parameters $\theta,w$
			\Ensure Optimal policy parameters $\theta^*$
			\While{not converged}
			\State Collect data $X=\left\{(s_i,a_i,r_i)\right\}_{i=1}^{N}$ using the current policy network $\pi_{\theta}$
			\State Use GAE\cite{gae} technique to estimate advantages
			\For{each training  epoch}
			\For{each mini-batch}
			\State Use $X$ to update parameters $\theta,w$ by $(\ref{ppo-clip})$
			\EndFor
			\State Dropout strategy in $(\ref{dropout})$: $X\leftarrow\mathcal{D}(X)$
			\EndFor
			\EndWhile
		\end{algorithmic}  
	\end{algorithm}
	
	\begin{table}[h]
		\centering
		\caption{Detailed hyperparameters} \label{Detailed hyperparameters}
		\begin{tabular}{c|cc}
			\toprule
			Algorithm & PPO\cite{ppo} & D-PPO (\textcolor{red}{ours}) \\
			\midrule
			Number of actors & 8 & 8 \\
			Horizon & 256 & 256 \\
			Learning rate & $2.5\times10^{-4}$ & $2.5\times10^{-4}$ \\
			Learning rate decay & Linear & Linear \\
			Optimizer & Adam & Adam \\
			Total steps & $1\times10^7$ & $1\times10^7$ \\
			Batch size & 2048 & 2048 \\
			Update epochs & 4 & 4 \\
			Mini-batch size & 512 &  512 \\
			Mini-batches & 4 & 4 \\
			GAE parameter $\lambda$ & 0.95 & 0.95 \\
			Discount factor $\gamma$ & 0.99 & 0.99 \\
			Clipping parameter $\epsilon$ & 0.1 & 0.1\\
			Value loss coeff. $c_1$ & 1 & 1 \\
			Entropy loss coeff. $c_2$ & 0.01 & 0.01 \\
			Dropout ratio $r$ & $\backslash$ & 0.2 \\
			\bottomrule
		\end{tabular}
	\end{table}


	\begin{figure*}[!t]
		\centering
		
		\subfloat[Boxing]{
			\includegraphics[scale=0.57]{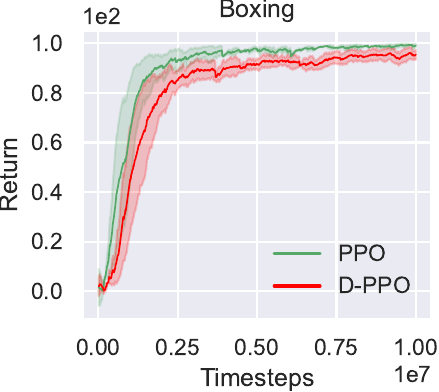}
			\includegraphics[scale=0.57]{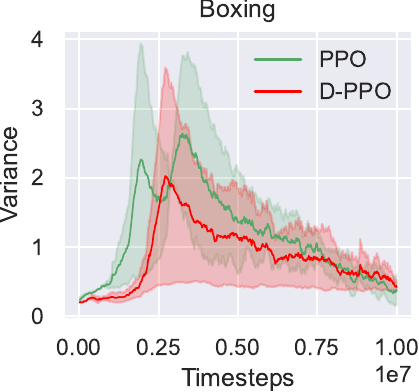}
		}
		\subfloat[Breakout]{
			\includegraphics[scale=0.57]{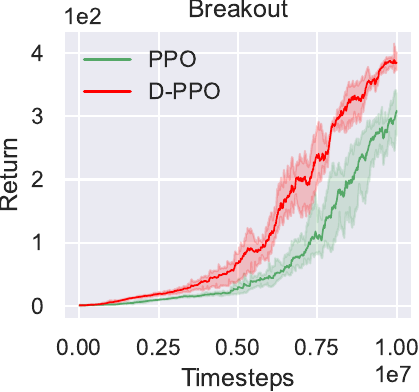}
			\includegraphics[scale=0.57]{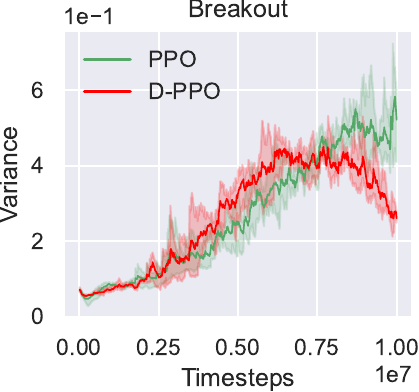}
		}
		
		\subfloat[CrazyClimber]{
			\includegraphics[scale=0.57]{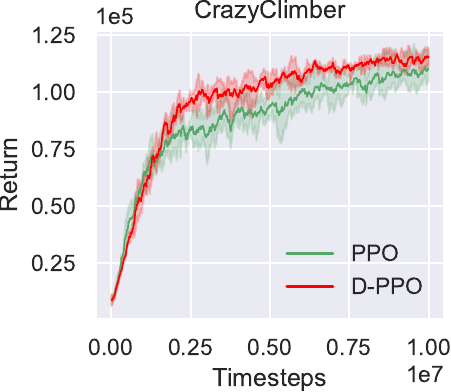}
			\includegraphics[scale=0.57]{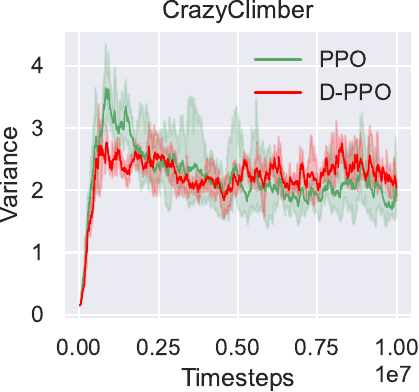}
		}
		\subfloat[DemonAttack]{
			\includegraphics[scale=0.57]{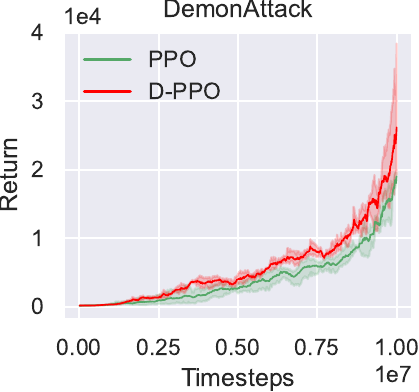}
			\includegraphics[scale=0.57]{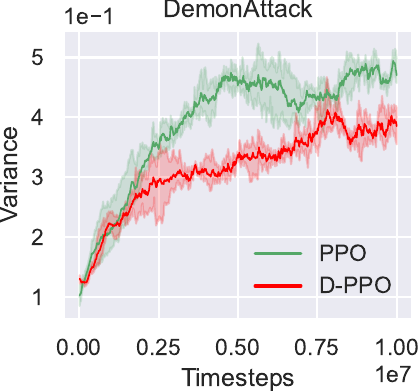}
		}
		
		\subfloat[Enduro]{
			\includegraphics[scale=0.57]{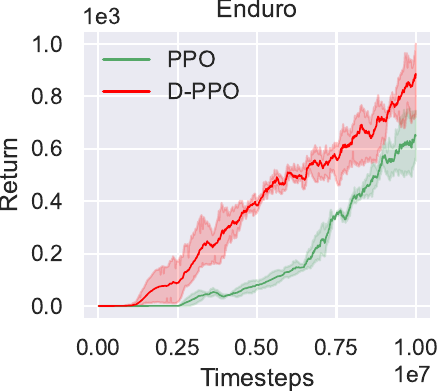}
			\includegraphics[scale=0.57]{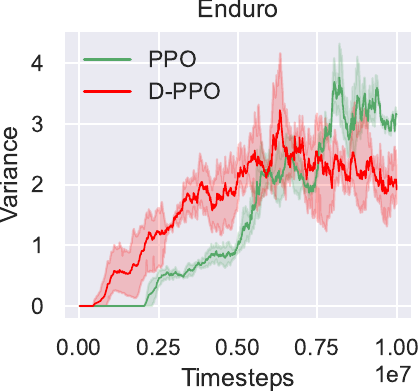}
		}
		\subfloat[Gravitar]{
			\includegraphics[scale=0.57]{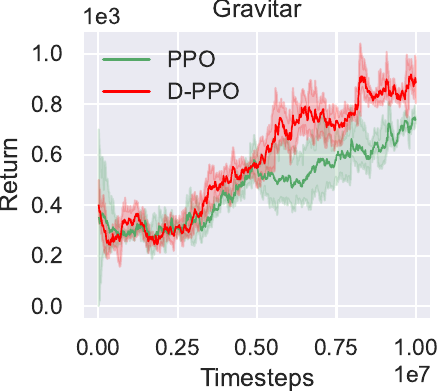}
			\includegraphics[scale=0.57]{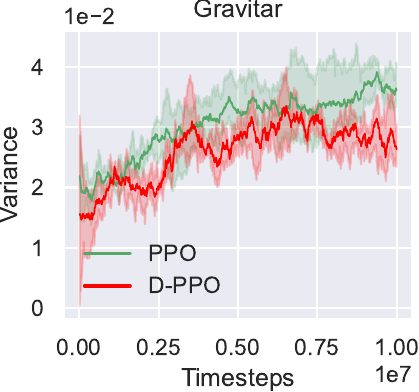}
		}
		
		\subfloat[Kangaroo]{
			\includegraphics[scale=0.57]{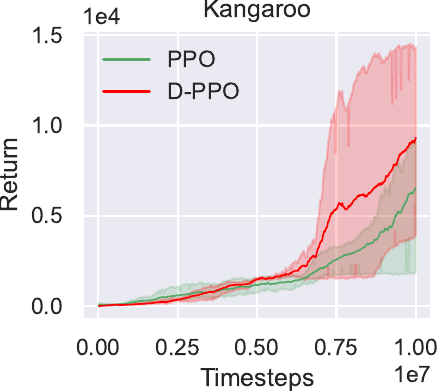}
			\includegraphics[scale=0.57]{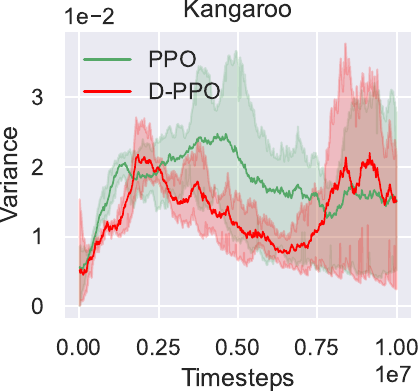}
		}
		\subfloat[SpaceInvaders]{
			\includegraphics[scale=0.57]{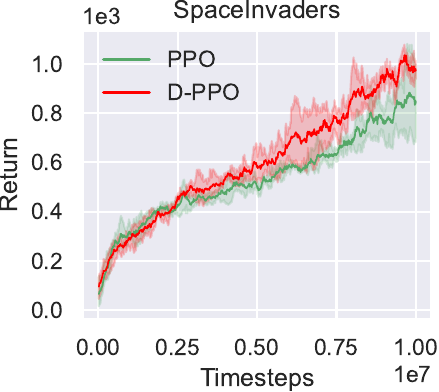}
			\includegraphics[scale=0.57]{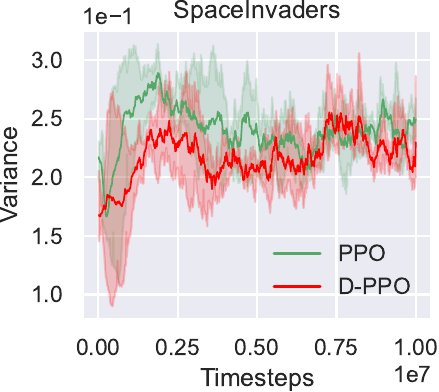}
		}
		\caption{The training curves (left) and surrogate objective variances (right) for PPO and D-PPO algorithms in different environments (five sets of experiments repeated for each environment with different random seeds).}
		\label{result_1}
	\end{figure*}

	\begin{table*}[!t]
		\centering
		\caption{Average return of all training steps.}\label{Average return of all time steps}
		\begin{tabular}{c|cccccccc}
			\toprule
			Environment& Boxing  &Breakout &CrazyClimber  & DemonAttack &Enduro &Gravitar&Kangaroo & SpaceInvaders  \\
			\midrule
			PPO & \bfseries90.923&79.615 & 90283.967&  4426.82&194.552 &477.483&1824.8&549.283\\
			D-PPO &83.171&\bfseries135.457 & \bfseries97928.3&  \bfseries6184.817&\bfseries391.222 &\bfseries579.433&\bfseries2792.167&\bfseries618.807\\
			Improvement & $-7.752$&\textcolor{red}{$+55.842$} & \textcolor{red}{$+7644.333$}&  \textcolor{red}{$+1757.997$}&\textcolor{red}{$+196.671$} &\textcolor{red}{$+101.94$}&\textcolor{red}{$+967.366$}&\textcolor{red}{$+69.523$}\\
			Improvement ($\%$)&$-8.526\%$&\textcolor{red}{$+70.139\%$} & \textcolor{red}{$+8.467\%$}& \textcolor{red}{$+39.712\%$}&\textcolor{red}{$+101.089\%$} &\textcolor{red}{$+21.352\%$}&\textcolor{red}{$+53.012\%$}&\textcolor{red}{$+12.657\%$}\\
			\bottomrule
		\end{tabular}
	\end{table*}

	\begin{table*}[!t]
		\centering
		\caption{Average return for the last 0.1 million training steps}\label{Average return 2}
		\begin{tabular}{c|cccccccc}
			\toprule
			Environment&  Boxing  &Breakout &CrazyClimber  & DemonAttack &Enduro &Gravitar&Kangaroo & SpaceInvaders  \\
			\midrule
			PPO  &\textbf{99.6} &328.733 &108486.667 &21778.667 &721.9 &735.0 &6860.0&825.333\\
			D-PPO  &95.8 &\textbf{380.267} &\textbf{115530.0} &\textbf{29653.0} &\textbf{944.067} &\bfseries860.0 &\bfseries9473.333&\textbf{973.667}\\
			Improvement & $-3.8$&\textcolor{red}{$+51.533$} &\textcolor{red}{$+7043.333$} &\textcolor{red}{$+7874.333$} &\textcolor{red}{$+222.167$}&\textcolor{red}{$+125.0$}&\textcolor{red}{$+2613.333$}&\textcolor{red}{$+148.333$} \\
			Improvement ($\%$)&$-3.815\%$ &\textcolor{red}{$+15.676\%$} &\textcolor{red}{$+6.492\%$} &\textcolor{red}{$+36.156\%$} &\textcolor{red}{$+30.775\%$}&\textcolor{red}{$+17.007\%$}&\textcolor{red}{$+38.095\%$}&\textcolor{red}{$+17.973\%$} \\
			\bottomrule
		\end{tabular}
	\end{table*}

	\begin{table*}[!t]
		\centering
		\caption{Average return for the last 1 million training steps}\label{Average return 3}
		\begin{tabular}{c|cccccccc}
			\toprule
			Environment& Boxing  &Breakout &CrazyClimber  & DemonAttack &Enduro  &Gravitar&Kangaroo& SpaceInvaders  \\
			\midrule
			PPO &\textbf{99.105} &276.495 &104950.0 &14780.775 &610.99 &715.833 &5741.333&841.833\\
			D-PPO&95.185 &\textbf{379.61} &\textbf{113531.0} &\textbf{21224.875} &\textbf{811.08} &\bfseries865.0&\bfseries8674.666&\textbf{981.75} \\
			Improvement&$-3.92$ &\textcolor{red}{$+103.115$} &\textcolor{red}{$+8581.0$} &\textcolor{red}{$+6444.1$} &\textcolor{red}{$+200.09$} &\textcolor{red}{$+149.166$}&\textcolor{red}{$+2933.333$}&\textcolor{red}{$+139.917$} \\
			Improvement ($\%$)&$-3.955\%$ &\textcolor{red}{$+37.294\%$} &\textcolor{red}{$+8.176\%$} &\textcolor{red}{$+43.598\%$} &\textcolor{red}{$+32.748\%$} &\textcolor{red}{$+20.838\%$}&\textcolor{red}{$+51.092\%$}&\textcolor{red}{$+16.620\%$} \\
			\bottomrule
		\end{tabular}
	\end{table*}

	\begin{figure*}[!t]
		\centering
		\includegraphics[scale=0.7]{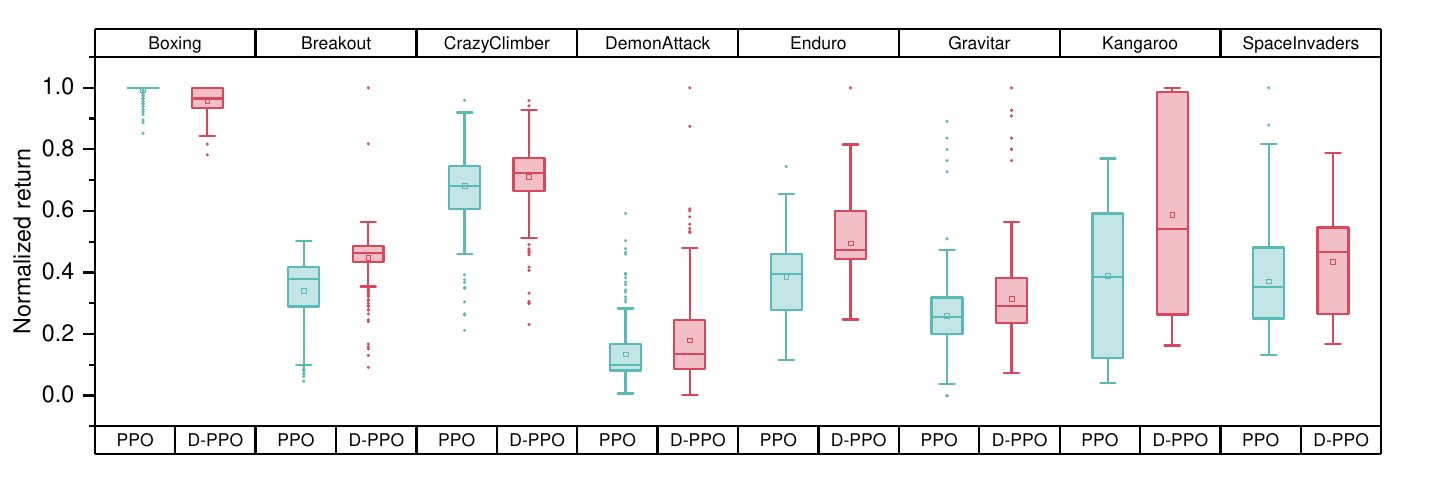}
		\caption{The normalized returns of PPO and D-PPO algorithms in different environments during the last 1 million training steps (five sets of experiments repeated for each environment with different random seeds).}\label{box}
	\end{figure*}

	\subsection{Comparative Experiments}
	Our policy network and value network structure are shown in Fig. \ref{network structure}. The input to the network is the stacking of the last four frames resized to 84 x 84 x 4. The value network and policy network share the same convolutional layer to improve learning efficiency. The network structure and common hyperparameters of the PPO and D-PPO algorithms are completely consistent. Specifically, the learning rate is set to $2.4\times10^{-4}$ and linearly decreases to 0, and the total number of steps of interaction with the environment is $1\times10^7$. We have eight intelligent agents that share the latest parameters and interact independently with the environment. The batch size is set to 2048, and each round of updates trains for 4 epochs, with each epoch divided into 4 mini-batches for updates. We used the GAE\cite{gae} technique to estimate the advantage function, with related parameters $\lambda$ and $\gamma$ set to 0.95 and 0.99. Our final loss consists of three parts, that is, 
	\begin{equation}
		l=l_p+c_1\cdot l_v-c_2\cdot l_e,
	\end{equation}
	where $l_p$ and $l_v$ are the losses of policy and value network, $l_e$ is the entropy of policy network output, the weight coefficients are set as $c_1=1$ and $c_2=0.01$. The clipping ratio $\epsilon$ is set to 0.1, and the dropout ratio $r$ of D-PPO is set to 0.2, as shown in Table \ref{Detailed hyperparameters}.
	
	The experimental results are shown in Fig. \ref{result_1}. It can be seen that, except for Boxing environment, the performance of D-PPO algorithm is slightly lower than that of PPO algorithm. However, in all other environments, there is a certain performance improvement, especially in Breakout, Enduro, Gravitar and Kangaroo, where there is a significant performance improvement.

	For the variance of the surrogate objective, It can be seen that the D-PPO algorithm can effectively limit the surrogate objective variance in almost all environments except for the CrazyClimber. In the Breakout environment, the return of the D-PPO algorithm is much higher than that of the PPO algorithm before about 7 million steps, which also leads to a larger surrogate objective variance compared to the PPO algorithm. After 7 million steps, with the decrease of the learning rate, the surrogate objective variance of the D-PPO algorithm decreases gradually due to the dropout strategy and is significantly lower than that of the PPO algorithm. This experimental phenomenon is also very evident in the Enduro environment, after approximately 7.5 million steps, the surrogate objective variance of the D-PPO algorithm can be observed to gradually decrease and smaller than that of the PPO algorithm. In the DemonAttack and Gravitar environments, the effectiveness of the D-PPO algorithm is even more evident: its surrogate objective variance is lower than that of the PPO algorithm at almost all time steps.
	
	In addition, Tables \ref{Average return of all time steps}, \ref{Average return 2} and \ref{Average return 3} show the average returns of PPO and D-PPO algorithms for all time steps, for the last 0.1 million steps and for the last 1 million steps in the corresponding environments under five random seeds. It can be seen that the D-PPO algorithm has achieved stable performance improvements in all environments except for Boxing. Fig. \ref{box} shows the box plot of all the returns of PPO and D-PPO algorithms in the last 1 million steps in the corresponding environments under five random number seeds. It can be seen that significant performance improvement was achieved in Breakout, CrazyClimber, and Enduro environments, and the returns had smaller variance.
	
	In general, from the experimental results, the performance of the D-PPO algorithm is superior in most environments, and it can effectively limit the excessive growth of the variance of the surrogate objective, which is a direct proof of the effectiveness of the dropout strategy.

	\begin{figure*}[!h]
		\centering
		\subfloat[Breakout]{
			\includegraphics[scale=0.57]{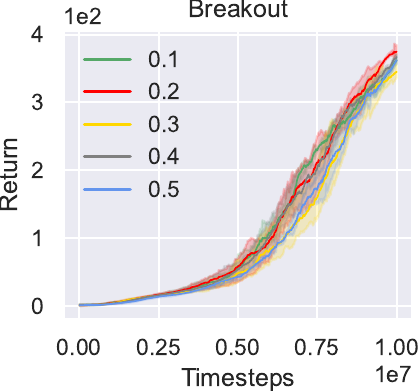}
			\includegraphics[scale=0.57]{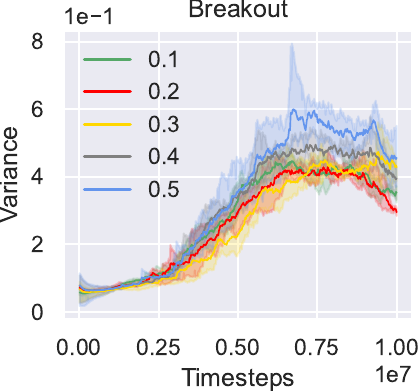}
			\includegraphics[scale=0.57]{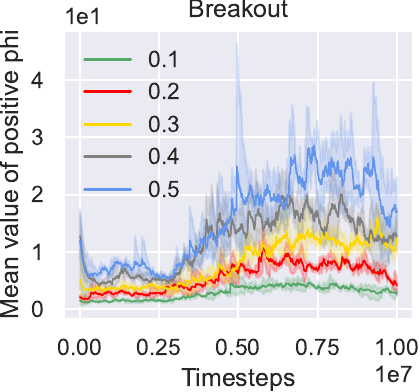}
			\includegraphics[scale=0.57]{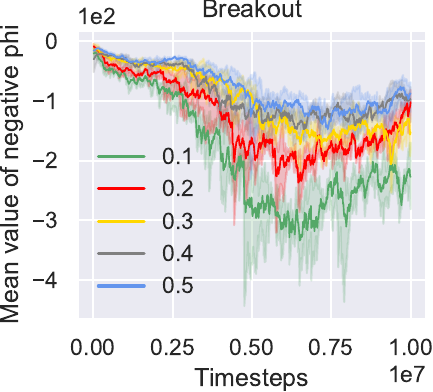}
		}
		
		\subfloat[Enduro]{
			\includegraphics[scale=0.57]{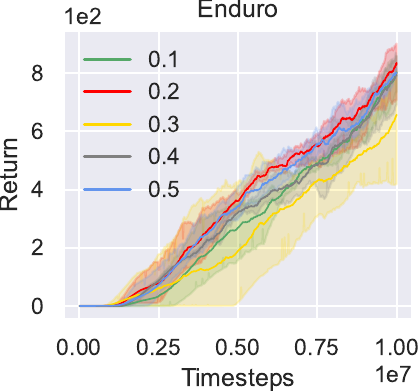}
			\includegraphics[scale=0.57]{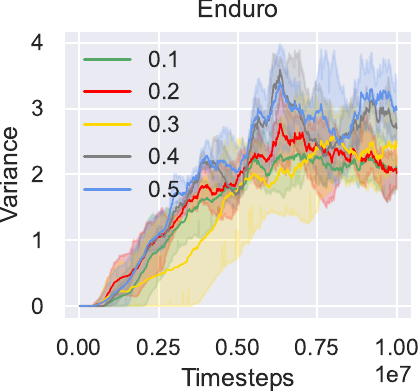}
			\includegraphics[scale=0.57]{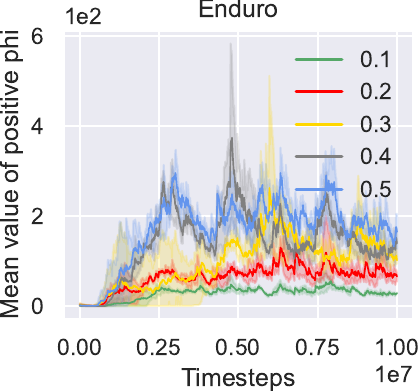}
			\includegraphics[scale=0.57]{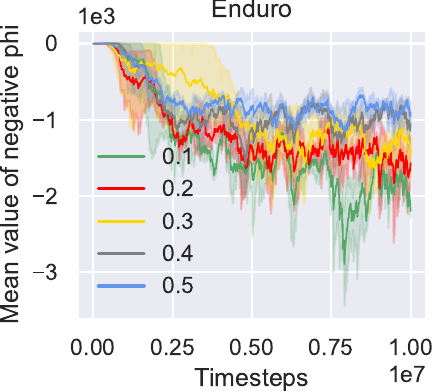}
		}
		
		\subfloat[SpaceInvaders]{
			\includegraphics[scale=0.57]{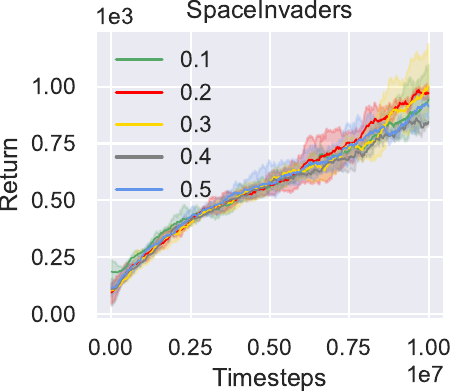}
			\includegraphics[scale=0.57]{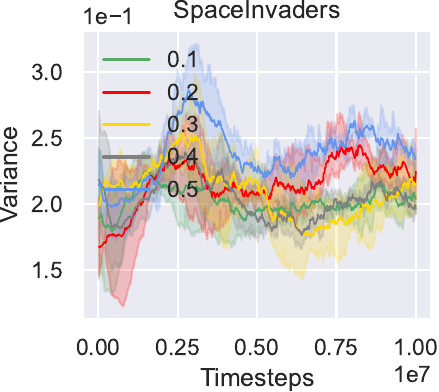}
			\includegraphics[scale=0.57]{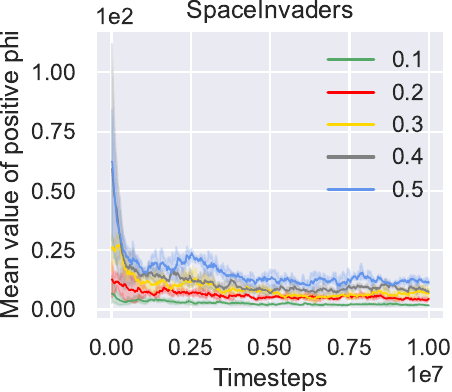}
			\includegraphics[scale=0.57]{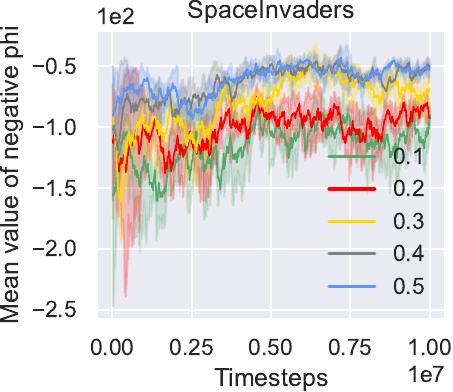}
		}
		\caption{The training curves corresponding to different values of $r$ in the D-PPO algorithm under different environments. They are returns, the variances of the surrogate objective, and the average values of $\phi(x)$ that is positive and negative in the dropout data, respectively (five sets of experiments repeated for each environment with different random seeds).}
		\label{hyperparameter_1}
	\end{figure*}

	\begin{figure*}[!t]
		\centering
		\subfloat[Breakout]{
			\includegraphics[scale=0.32]{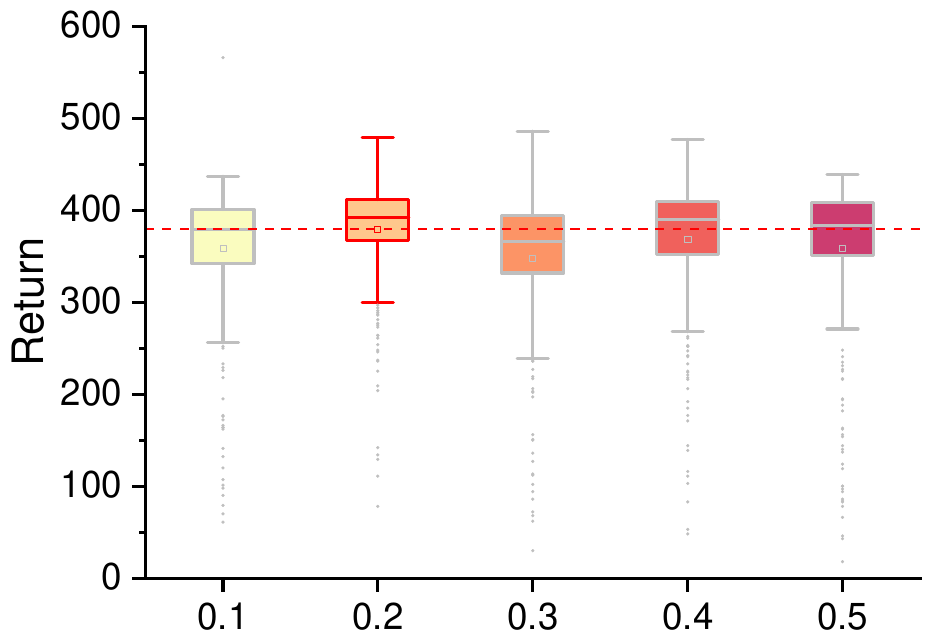}
		}
		\subfloat[Enduro]{
			\includegraphics[scale=0.32]{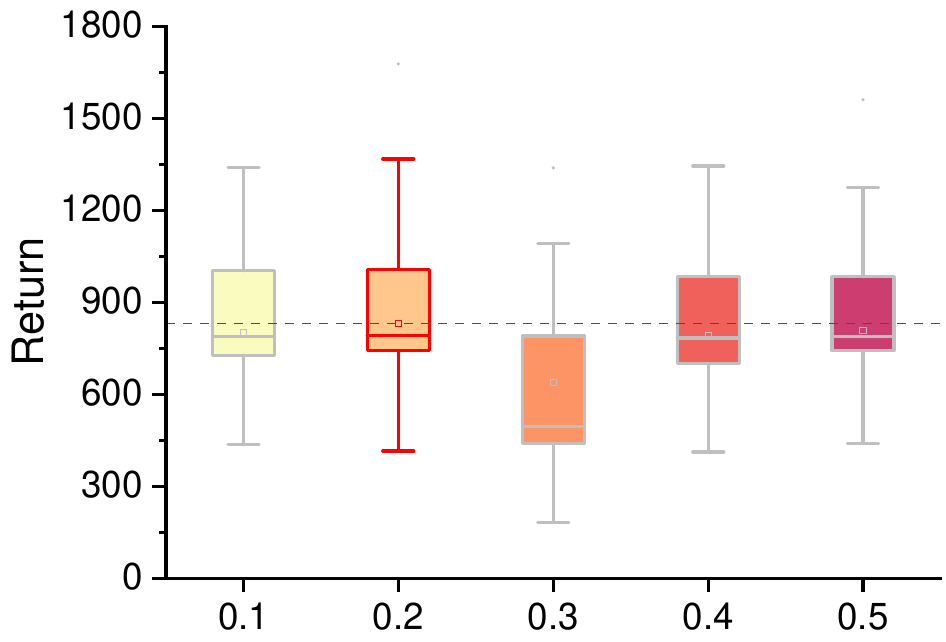}
		}
		\subfloat[SpaceInvaders]{
			\includegraphics[scale=0.32]{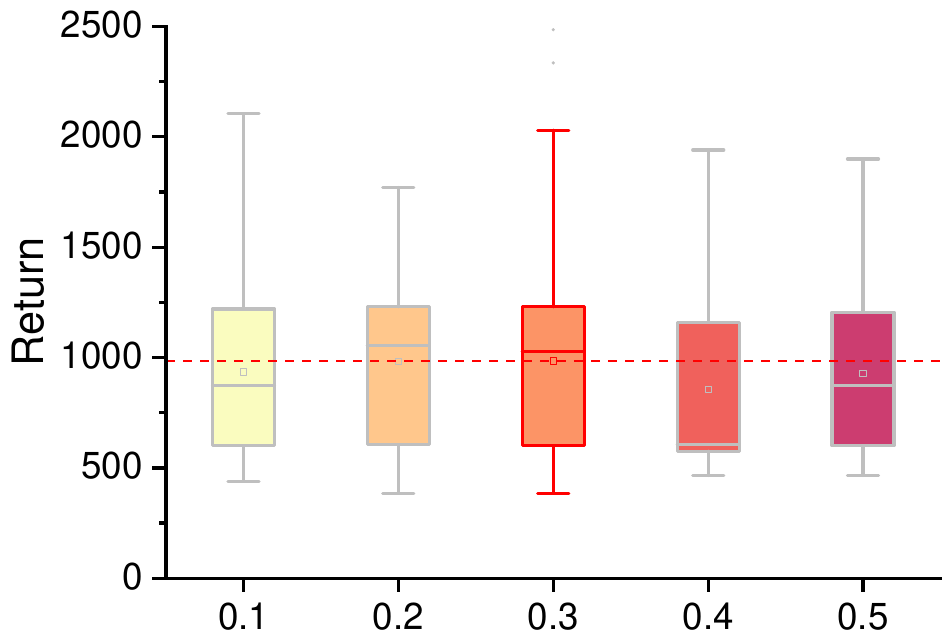}
		}
		\caption{The box plot of the returns for the last 1 million training steps corresponding to different values of $r$ in the D-PPO algorithm under three environments, here the box with the largest mean value highlighted in red (five sets of experiments repeated for each environment with different random seeds).}
		\label{hyperparameter_2}
	\end{figure*}

	\subsection{Hyperparameter Analysis}
	Our main question now is how to set the hyperparameter $r$ in the D-PPO algorithm? To answer this question, in this section we set comparative experiments with different hyperparameters $r$ to determine the optimal one.

	We selected three representative environments, namely Breakout, Enduro, and SpaceInvaders, and conducted repeated experiments on five different random seeds. The experimental results are shown in Fig. \ref{hyperparameter_1} and Fig. \ref{hyperparameter_2}. From the perspective of return, Fig. \ref{hyperparameter_2} and the first column of Fig. \ref{hyperparameter_1} reflect the returns of D-PPO algorithm under different hyperparameters $r\in\left\{0.1,0.2,0.3,0.4,0.5\right\}$. It can be seen that D-PPO algorithm achieves the highest average return when $r=0.2$ in Breakout and Enduro environments. From the perspective of surrogate objective variance, as shown in the second column of Fig. \ref{hyperparameter_1}, when $r$ is set to 0.1 or 0.2, D-PPO algorithm effectively limits the growth of surrogate objective variance in Breakout and Enduro environments. In addition, the third and fourth columns of Fig. \ref{hyperparameter_1} respectively represent the average values of $\phi(x)$ that is positive and negative in the dropout data. It can be seen that as $r$ increases, its value gradually increases, which is intuitive and also indicates the rationality of the dropout strategy. Therefore, based on the above analysis, we recommend setting the hyperparameter $r$ of the D-PPO algorithm to 0.2, as it achieves the highest average return in multiple environments and is able to more effectively limit the variance of the surrogate objective.

	\section{Conclusion}\label{Conclusion}
	In this article, a dropout strategy framework for policy optimization methods was proposed. Under this framework, we derive an upper bound on the variance of the surrogate objective, and propose a dropout strategy to limit the excessive growth of the surrogate objective variance caused by the introduction of importance sampling. By applying the dropout strategy to the PPO algorithm, we obtain the D-PPO algorithm. We conducted comparative experiments between the D-PPO and PPO algorithms in the Atari 2600 environment to verify the effectiveness of the dropout strategy, and further discussed the setting of the hyperparameter $r$ in D-PPO. 
	
	There is still space for improvement. The dropout strategy may also pose risks to policy optimization algorithms, as discarding some sample data to reduce the variance of the surrogate objective may also result in the dropout of important samples that can significantly improve the performance of the policy network. An interesting direction for future work is to apply the dropout strategy to a wider range of policy optimization methods and simulation environments, and try to avoid the above situations, which we will consider as our research goal in the next stage.

\end{document}